# Utilities as Random Variables: Density Estimation and Structure Discovery

**Urszula Chajewska**
Computer Science Department
Stanford University
Stanford, CA 94305-9010
*urszula@cs.stanford.edu*

**Daphne Koller**
Computer Science Department
Stanford University
Stanford, CA 94305-9010
*koller@cs.stanford.edu*

## Abstract

Decision theory does not traditionally include uncertainty over utility functions. We argue that the a person's utility value for a given outcome can be treated as we treat other domain attributes: as a random variable with a density function over its possible values. We show that we can apply statistical density estimation techniques to learn such a density function from a database of partially elicited utility functions. In particular, we define a Bayesian learning framework for this problem, assuming the distribution over utilities is a mixture of Gaussians, where the mixture components represent statistically coherent subpopulations. We can also extend our techniques to the problem of discovering generalized additivity structure in the utility functions in the population. We define a Bayesian model selection criterion for utility function structure and a search procedure over structures. The factorization of the utilities in the learned model, and the generalization obtained from density estimation, allows us to provide robust estimates of utilities using a significantly smaller number of utility elicitation questions. We experiment with our technique on synthetic utility data and on a real database of utility functions in the domain of prenatal diagnosis.

## 1 Introduction

The principle of maximizing expected utility has long been established as the guide to making rational decisions [21]. It rests on two components: probabilities for representing our uncertainty about the situation, and utilities for representing our preferences.

Traditional decision theory ignores, however, any uncertainty we may have about the utilities of a given user. To apply it, we need to acquire the entire utility function. We cannot use any prior knowledge, either in the form of experience with other users or in the form of constraints. By treating utilities as random variables, we can utilize tools that have been used with great success when reasoning about events in decision problems. For example, we can use value of information to decide whether a utility elicitation question is worth asking [4].

Before we can apply these tools, however, we need to address the issue of acquiring distributions over utilities. The problem of model acquisition is well-understood in the context of probabilistic models, with a significant body of work both on eliciting models from experts and on learning from sample data. By contrast, the problem of acquiring utility functions is not understood nearly as well. In some sense, utility elicitation is innately harder. There are no experts to ask about the model; every person's utility function may be different. Thus, in the traditional approach, each individual's utility for each of the possible outcomes must be elicited. In domains involving more than a few outcomes, this elicitation process is time consuming and cognitively difficult. It is also noisy and prone to errors [15].

The use of structure is crucial for probabilities, simplifying both the model and the associated knowledge acquisition process. Structure also exists in utilities. Utility functions can often be decomposed as a linear combination of *subutility functions*, each of which involves only a few of the relevant variables. Decomposable utility functions can be used to support more efficient inference [14, 20]. In principle, as they require fewer parameters to be specified, they should also ease the knowledge acquisition process [15].

In practice, however, we see that decomposable utility functions are rarely used. (Except in certain settings where everything easily reduces to a common basis, such as money.) The problem is that the structure in utility functions seems elusive, perhaps because there is little methodology for discovering it. Several papers [9, 17] have tried to detect simple additive decompositions in a database of elicited utility functions using linear regression; unfortunately, additive structure rarely seems to exist in these databases, so one typically resorts back to explicit utility elicitation for the entire outcome space. We know of no attempts to learn more complex utility functions from data. Alternatively, one could ask specific individuals about their decomposition. However, this approach is difficult to implement. Unlike probabilities, utilities cannot be marginalized. The utility of a specific instantiation of one state attribute does not have any intuitive meaning and cannot be assessed without making some assumptions about the values of other attributes. Thus, the decomposition of utility functions is much less intuitive for people to understand than the decomposition of probability functions.

In this paper, we take a much more general approach to the problem of discovering the structure of utility functions. We assume that we have access to a database of (partially) elicited utility functions for some set of individuals. This assumption is not unreasonable: many medical informatics centers collect large databases of utility functions for various decision problems or for cost-benefit analyses of new treatments [10, 18]. Given such a database, we apply statistical learning techniques to discover a decomposition that fits the data well. More specifically, we postulate a model where the population is comprised of several statistically coherent subpopulations, or clusters. The utility functions in each cluster are assumed to be decomposed in some way, and the parameters of the subutilities are assumed to come from a Gaussian distribution. Note that we do not assume that any of these model parameters are known. We do not know which person belongs to which cluster, or even which decomposition is used in the different clusters. Rather, we are given only a standard database of fully explicit utility functions (where some of the values may be missing).

Our approach allows us to learn substantially more expressive models than the naive linear regression approach, and thereby discovers structures that are invisible to linear regression. Furthermore, the model produced by our learning algorithm can be used to make the utility elicitation process more robust and easier for the user.

## 2 Factored Utility Functions

The naive representation of a utility function is a vector of real numbers, ascribing a utility to each possible outcome. This representation is quite reasonable in domains involving a small number of distinct outcomes. Many real-life domains, however, involve fairly complex outcomes. Consider, for example, the domain of prenatal testing. Prenatal testing is intended to diagnose the presence of a chromosomal abnormality such as Down's syndrome in the early weeks of pregnancy, an event whose probability increases with maternal age. The two tests currently available to diagnose it, chorionic villus sampling (CVS) and amniocentesis (AMNIO), carry a significant risk of miscarriage above the baseline rate. The risk is higher for CVS, but it is more accurate and can be performed earlier in the pregnancy. Both miscarriage and elective termination of the pregnancy may reduce the chances of future pregnancy. In this domain, a typical outcome is "healthy fetus, early test (CVS), accurate test result, procedure-related miscarriage, no future pregnancy".

In such cases, it is convenient to describe the space of outcomes as the set of possible assignments of values to a set of relevant variables. Here, we have five utility attributes: testing $T$ (none, CVS or amniocentesis), fetus's status $D$ (normal, affected by Down's syndrome), possible loss of pregnancy $L$ (no loss, miscarriage, elective termi-nation), knowledge of the fetus's status $K$ (none, accurate, inaccurate), and future successful pregnancy $F$ (true, false). The utility is a function of all of these values. In general, we define each outcome as an assignment to a set of attribute variables $\mathbf{X} = \{X_1, \ldots, X_n\}$. Each variable $X_i$ has a domain $\text{Dom}(X_i)$ of two or more elements.

Clearly, the number of outcomes is exponential in the number of attributes. Thus, the specification of the utility function in full can become expensive. In many medical domains, there are tens of outcomes. In our domain, there are 108 distinct outcomes; even after simplification and elimination of very unlikely outcomes, 66 outcomes remain. Utility elicitation, which in the best of cases is a long and tiring process, is extremely difficult for outcome spaces of this size.[1]

In many cases, however, the utility function is not a single amorphous function over the space of outcomes. Rather, it exhibits some structure. One particularly important subclass of utility functions are those that decompose into components associated with smaller sets of attributes. For example, in a vacation planning domain, we might be able to construct our overall utility as a sum of functions associated with the cost of the vacation, with the weather in our destination, with the quality of the accommodations, etc. This type of decomposition lies at the heart of *multi-attribute utility theory* [15].

**Definition 2.1:** Let $C$ be a set of clusters of variables $\mathbf{C}_1, \ldots, \mathbf{C}_r$. We say that a utility function is *factored according to* $C$ if there exist functions $u_i : \text{Dom}(\mathbf{C}_i) \mapsto \mathbb{R}$ ($i = 1, \ldots, r$) such that $u(\mathbf{x}) = \sum_i u_i(\mathbf{c}_i)$ where $\mathbf{c}_i$ is the assignment to the variables in $\mathbf{C}_i$ in $\mathbf{x}$. We call the functions $u_i$ *subutility functions*. ∎

The factorization of the utility function induces observable patterns for the utilities of related outcomes. Some of these cases have received a lot of attention in the literature. For example, if the clusters are disjoint, then the change in the utility resulting from changing the assignment to the variables in one cluster does not depend on the assignments to the variables in the other clusters. In this case, the utility function is said to be *additive* over $C$. The intuitive behavior induced by additive utility functions makes them relatively easy to describe to a user and to test for during the process of utility elicitation.

A related concept is that of *conditionally additive* utility functions. Let $\mathbf{Y}, \mathbf{Z}, \mathbf{V}$ be a disjoint partition of $\mathbf{X}$. We say that $\mathbf{Y}$ and $\mathbf{V}$ are *conditionally additively independent given* $\mathbf{Z}$ if, for any fixed value $\mathbf{z}$ of $\mathbf{Z}$, we have that $\mathbf{Y}$ and $\mathbf{V}$ are additively independent in the utility function $u(\mathbf{Y}, \mathbf{V}, \mathbf{z})$. This type of decomposition is also relatively easy to test for, and hence is usable.

---

[1] In this prenatal testing domain, the speed of utility elicitation was around 10 outcomes per hour [16]. We were also told by several utility elicitation practitioners that the probability of inconsistent answers rises sharply after the first few questions as the fatigue grows.

However, the definition of factored utility functions covers many more cases than these special cases. Consider, for example, a set of clusters $C$ consisting of the three clusters $\{A,B\}, \{B,C\}, \{C,A\}$. As pointed out by Bacchus and Grove [1], a utility function that factorizes in this way does not have any of the commonly defined additive independence properties. They call such models *generalized additively independent*. They continue to say that, while utility functions that factorize in this way may well be useful in practice, their lack of intuitive semantics makes them hard to incorporate into a utility elicitation process.

Factored utility functions can be incorporated very naturally into influence diagrams [13]. Moreover, a factored utility function can be exploited by standard clique tree inference algorithms to make decision making more efficient, in much the same way as factored probability distributions are exploited in Bayesian network inference [14, 20].

Factored utilities admit a representation in terms of subutility functions over a much smaller domain. They can therefore be specified using a much smaller set of parameters. However, there are many slightly different ways to parameterize a factored utility function over $C$. We choose one that will allow us to make our learning algorithm more efficient.

**Definition 2.2:** We say that two functions $h, h'$ over some domain $\Omega$ are *orthogonal* if $\sum_{\omega \in \Omega} h(\omega) \cdot h'(\omega) = 0$. ∎

Our goal will be to construct a fixed *basis* $h_C$ of orthogonal functions, and represent a factored utility function $u$ over $C$ as a linear combination of the functions in this basis. The coefficients $\mathbf{w}$ of the different basis functions would be the parameters specifying $u$. The orthogonality property will allow us to perform the computation described in the subsequent sections more efficiently.

The atomic units in the construction of our basis are the basis functions that depend only on a single variable. For each variable $X$ with values $x_1, \ldots, x_k$, we define a set of $k$ basis functions $h_1^X, \ldots, h_k^X : \text{Dom}(X) \mapsto \mathbb{R}$. Our construction is such that:

- $h_1^X \equiv 1$, i.e., $h_1^X(x_i) = 1$ for all $i$;
- the $h_i^X$ functions are pairwise orthogonal.

For a binary-valued attribute $B$, we simply define:

$$h_2^B(x_1) = 1$$
$$h_2^B(x_2) = -1$$

For a three-valued attribute $C$, we define:

$$\begin{array}{rcl|rcl} h_2^C(x_1) &=& 1 & h_3^C(x_1) &=& 1 \\ h_2^C(x_1) &=& 0 & h_3^C(x_1) &=& -2 \\ h_2^C(x_1) &=& -1 & h_3^C(x_1) &=& 1 \end{array}$$

In general, we can define a set $\mathcal{H}[X]$ of orthogonal basis functions for any $k$-ary variable $X$. Note that, as the functions are orthogonal, they span the space of all possible functions over $X$. In other words, for every function $u : X \mapsto \mathbb{R}$, there exist coefficients $w_1, \ldots, w_k$ such that $u = \sum_{i=1}^{k} w_i h_i^X$.

We now use these basic building blocks to construct an orthogonal basis for functions over the entire set of outcomes. With a slight abuse of notation, we will view a function $h_i^X$ as a function over $\text{Dom}(\mathbf{V})$. Let $o$ be an outcome; recall that $o$ defines a value $X[o]$ for each variable $X \in \mathbf{V}$. We simply define $h_i^X(o) = h_i^X(X[o])$.

We can now define a basis for a cluster of variables $\mathbf{C}$ as the set of all functions that are composed as products of basis functions for the individual variables in $\mathbf{c}$:

$$\mathcal{H}[\mathbf{C}] = \{ \prod_{X \in \mathbf{C}} h^X : h^X \in \mathcal{H}[X] \}.$$

**Proposition 2.3:** *The functions in $\mathcal{H}[\mathbf{C}]$ are pairwise orthogonal, and the set $\mathcal{H}[\mathbf{C}]$ exactly spans the set of all possible functions over $\mathbf{C}$.*

By taking the union of the bases for the appropriate clusters, we can span any set of factored utility functions.

**Corollary 2.4:** *Let $C$ be a set of clusters. The set of functions $\mathcal{H}[C] = \cup_{\mathbf{c} \in C} \mathcal{H}[\mathbf{C}]$ spans the set of all factored utility functions over $C$.*

We can therefore parameterize any factored utility function over $C$ using a set of coefficients $w_i$, one for every function in $\mathcal{H}[C]$. How many parameters are required? For each cluster $\mathbf{C}$, we have $|\text{Dom}(\mathbf{C})|$ functions in $\mathcal{H}[\mathbf{C}]$. However, the bases for the different clusters are not disjoint.

**Example 2.5:** Assume that our clusters are $\{A\}, \{B,C\}$ and $\{C,D\}$, and that all of our variables are ternary. We have 3 functions in $\mathcal{H}[A]$, and 9 in each of $\mathcal{H}[\{B,C\}]$ and $\mathcal{H}[\{C,D\}]$. However, the $h_1$ (all 1) function is common to all clusters, and the three $h^C$ functions are common to the two clusters that contain $C$. Of course, we must be careful not to undercount by doublecounting the overlap: $h_1$ is also among the three functions in $\mathcal{H}[C]$. A careful count reveals that the total number of distinct functions in our basis is $3 + 9 + 9 - 3 - 1 - 1 + 1 = 17$. ∎

In general, we can compute the total number of distinct functions in our basis by a simple inclusion-exclusion formula, keeping in mind that the overlap between the bases for two clusters $\mathbf{C}$ and $\mathbf{C}'$ is precisely the basis for $\mathbf{C} \cap \mathbf{C}'$ (taking $\mathcal{H}[\emptyset]$ to be the single vector $h_1$):

$$\begin{aligned} |\mathcal{H}[C]| &= \sum_{i} |\mathcal{H}[\mathbf{C}_i]| - \sum_{i_1 \neq i_2} |\mathcal{H}[\mathbf{C}_{i_1} \cup \mathbf{C}_{i_2}]| \\ &= + \sum_{i_1 \neq i_2 \neq i_3} |\mathcal{H}[\mathbf{C}_{i_1} \cup \mathbf{C}_{i_2} \cup \mathbf{C}_{i_3}]| - \cdots \end{aligned}$$

Thus, the total number of basis functions, and thereby of parameters required, grows (at most) linearly with the number of clusters and exponentially with the size of each one.

## 3 The Basic Framework

Our approach relies on a few basic assumptions about the population of users whose utility we are trying to model. The first assumption is that the population is composed of several disjoint subpopulations, or types (which we model using a random variable $T$), where the utility functions of the individuals of each type are statistically similar. Each subpopulation may utilize a different factorization $C_t$ of the utility function. Thus, every individual is associated with a vector $\mathbf{w}_t$ of dimension $m_t = |\mathcal{H}[C_t]|$, where each $w_j$ is the coefficient of the $j$th basis function $h_j \in \mathcal{H}[C_t]$. The vector $\mathbf{w}_t[j]$ represents the user's subutility functions.

We represent a probabilistic model over utilities by defining a vector random variable $\mathbf{W}_t$. For each value $t$ of $T$, $P(\mathbf{W}_t \mid t)$ is a multivariate Gaussian with mean vector $\mu_t$ and covariance matrix $\Sigma_t$. We assume that individuals in the population are IID samples from the $P(\{\mathbf{W}_t\}_t \mid T)$ distribution.

An individual's subutility vector $\mathbf{w}_t$ defines a complete utility function, which specifies a utility for each of the $n = |\mathrm{Dom}(\mathbf{X})|$ outcomes $o$. We can define this implicit utility function using a simple matrix operation. Let $A_t$ be the $n \times m_t$ matrix $(a^t_{ij})$ where $a^t_{ij} = h_j(o_i)$ for $o_i$ the $i$th possible outcome. Then, the user's utility function ought to be $\mathbf{u}^* = A_t \mathbf{w}_t$. However, the utility elicitation process can be quite noisy. We accommodate for that by assuming that the user's actual utility vector $\mathbf{u}$ is modified by some white noise, i.e., for each $o$, we have that $u_o$ is $u_o^*$ plus some random white noise $\varepsilon_t$ sampled from a zero-mean Gaussian distribution with some variance $\sigma_t^2$. More formally, we have a vector random variable $\mathbf{U}$ of dimension $n$, which is a linear Gaussian whose mean is $A_t \mathbf{W}_t$ and whose variance is $\sigma_t^2 I$ where $I$ is the unit matrix.

Note that, for each type $t$, the distribution over $\mathbf{W}_t, \mathbf{U}$ is a simple multivariate Gaussian, defined using a Gaussian distribution over $\mathbf{W}_t$ and a conditional linear Gaussian for $\mathbf{U}$ given $\mathbf{W}_t$. However, the distribution as a whole is not exactly a mixture of linear Gaussians, as the dimension of the vector $\mathbf{w}_t$ varies for the different types.

A model such as this can be used for several purposes. The most basic use is to compute the most probable factored utility function for a given user. More precisely, assume we are given a vector $\mathbf{u}$ representing the full utility function elicited from a certain user. Our goal is to compute the type $t$ and vector $\mathbf{w}_t$ such that the probability $P(\mathbf{w}_t \mid \mathbf{u}, t)$ is maximized. We perform a separate computation for each $t$.

From the definition of our generative model, we have that: $P(\mathbf{w}_t \mid \mathbf{u}, t) = \frac{P(\mathbf{u} \mid \mathbf{w}_t) P(\mathbf{w}_t \mid t)}{P(\mathbf{u} \mid t)}$. The denominator is a constant, so it does not affect the choice of maximum. Furthermore, the individual components $U_o$ of the vector variable $\mathbf{U}$ are conditionally independent given $\mathbf{W}_t$, so that our goal is to maximize $(\prod_o P(u_o \mid \mathbf{W}_t)) \cdot P(\mathbf{W}_t \mid t)$. Maximizing this function is equivalent to minimizing an error function corresponding to its negative logarithm [2]:

$-\sum_o \ln P(u_o \mid \mathbf{w}_t) - \ln P(\mathbf{w}_t \mid t)$. The first term in our error function (for the given vector $\mathbf{u}$) can be simplified to

$$-\frac{1}{2\sigma_t^2} \sum_o ((A_t)_o \mathbf{w}_t - u_o)^2 + n \ln \sigma_t + \frac{n}{2} \ln(2\pi) \quad (1)$$

where $(A_t)_o$ is the row of the matrix $A_t$ that corresponds to the outcome $o$. Simplifying $-\ln P(\mathbf{w}_t \mid t)$, we get:

$$\frac{m_t}{2}\ln(2\pi) + \frac{1}{2}\ln|\Sigma_t| + \frac{1}{2}(\mathbf{W}_t - \mu_t)^T \Sigma_t^{-1} (\mathbf{W}_t - \mu_t). \quad (2)$$

If we put together (1) and (2), and eliminate terms that do not depend on $\mathbf{w}_t$, we get as our final error function:

$$\begin{aligned} E(\mathbf{w}_t) &= \frac{1}{2\sigma_t^2} \sum_o (A_t(o)\mathbf{w}_t - u_o)^2 + \frac{1}{2}(\mathbf{w}_t - \mu_t)^T \Sigma_t^{-1}(\mathbf{w}_t - \mu_t) \\ &= \frac{1}{2\sigma_t^2} \|A_t \mathbf{w}_t - \mathbf{u}\|^2 + \frac{1}{2}\|B_t \mathbf{w}_t - B_t \mu_t\|^2 \end{aligned}$$

where $B_t^T B_t = \Sigma_t^{-1}$. (We are guaranteed that such a decomposition exists because the covariance matrix of a Gaussian is guaranteed to be positive definite.)

Thus, maximizing the posterior probability of the vector $\mathbf{w}_t$ is equivalent to minimizing a squared-error function. Let $D_t$ be the $(n+m_t) \times m_t$ matrix obtained by concatenating the matrices $\frac{1}{\sigma_t} A_t$ and $B_t$. We also define a vector $\mathbf{u}'$ of length $n + m_t$ defined by concatenating $\frac{1}{\sigma_t} \mathbf{u}$ and $B_t \mu_t$.

Note that we designed the matrix $A_t$ to guarantee that the columns of $D_t$ are linearly independent. Thus, we can compute the optimal solution to the least-squares problem by projection [19]:

$$\begin{aligned} \hat{\mathbf{w}}_t &= (D_t^T D_t)^{-1} D_t^T \mathbf{u}' \\ &= (\frac{1}{\sigma_t^2} A_t^T A_t + \Sigma_t^{-1})^{-1} D_t^T \mathbf{u}' \end{aligned}$$

The matrix $(\frac{1}{\sigma_t^2} A_t^T A_t + \Sigma_t^{-1})^{-1} D_t^T$ does not depend on $\mathbf{u}$, and can therefore be computed once and reused for every individual for whom we want to estimate $\mathbf{w}_t$.

This computation gives us, for each type $t$, the most likely vector $\mathbf{w}_t$ for the user given that he is in class $t$. We can now easily compute the most likely pair $(t, \mathbf{w}_t)$ for this user.

This model can also be used to give us more information. Recall that the conditional distribution on $\mathbf{W}_t, \mathbf{U}$ is a multivariate Gaussian distribution. At the cost of a little more work, we can compute the entire posterior distribution $P(\mathbf{W}_t \mid \mathbf{u}, t)$. The result would also be a Gaussian distribution, over the variables $\mathbf{W}_t$. The mean of this distribution would be precisely the vector $\hat{\mathbf{w}}_t$ computed above. The covariance matrix of the distribution could be used as an indicator for how confident we are in our estimate $\hat{\mathbf{w}}_t$. Clearly, there are situations where this information can be quite important, but it is not clear that it is always worth the computational overhead. On the other hand, unlike projection, this technique can be used even if some of the values in the original utility vector are missing.

## 4 Model Learning

In the previous section, we defined a statistical model of utilities in a population of users, and showed how it can be used to compute a factorization of an elicited utility function. We now move to tackling the problem of acquiring such a statistical model.

Our goal is to acquire this model from a database of utility functions elicited from a random population of users. Even if the utility function is factored, the utility elicitation process is typically done in terms of utilities of full outcomes. This is certainly the case if, as we assumed, the factorization of the utility function is unknown in advance. Thus, we assume that the training data we are given is a set of utility vectors $\mathbf{u}[j]$, one for each individual. We assume that some of the components of the utility vectors may be missing. The type variable $T$ and the corresponding decomposed utility vector $\mathbf{W}_t$ are unobserved in the training data.

Our key subroutine is the parameter estimation task for a given model. While we cannot use full Bayesian estimation in the presence of partially observable data, it will nevertheless be useful to view the model parameters as having a prior and a posterior. This perspective will be useful both for smoothing our numerical estimates and to provide the appropriate bias for the structure selection task.

Suppose that, for every value $t$ of the variable $T$, we have an $m_t$ dimensional multivariate Gaussian with an unknown mean vector $\mu_t$ and an unknown covariance matrix $\Sigma_t$. An appropriate conjugate prior over $\mu_t$ and $\Sigma_t$ is the *Normal-Wishart* [7]. We use a Normal-Wishart prior for the parameters of each of the type-specific Gaussian distributions over $\mathbf{W}_t$ (one for each type $t$) and for the parameters of the conditional Gaussian over the $U_o$ given $U^*(o) = A_t(o)\mathbf{W}_t$. We assume that the parameters $\theta_t$ representing the prior probability $P(T = t)$ are distributed with a Dirichlet distribution.

The main problem is that our data is only partially observable, rendering full Bayesian estimation infeasible. We therefore resort to finding the MAP parameter estimate using the *expectation-maximization* (EM) algorithm [8]. More precisely, we use our parameter prior to define a Gaussian prior distribution over $\mathbf{W}_t, U$. For each instance $j$ and each type $t$, we condition this distribution on $\mathbf{u}[j]$, and obtain a Gaussian posterior $P(\mathbf{W}_t[j] \mid t, \mathbf{u}[j])$. We use these Gaussian distributions to compute *expected sufficient statistics*: the expected empirical means and expected empirical covariances. These are used to update the Wishart priors, which then generate a new Gaussian prior distribution over $\mathbf{W}_t, U$. A similar update is done to the Dirichlet distribution over the types. The process iterates until convergence. We describe this process in detail in Appendix A.

Now, we consider the problem of finding a good structure. We focus on the problem of discovering the structure of the subutility functions within the clusters, and assume the number of clusters is given. (Our techniques easily extend to the more standard problem of discovering the number of clusters.) We apply Bayesian model selection to this task. More precisely, we define a discrete variable $S$ whose states $s$ correspond to possible models, i.e., possible decompositions of the subutilities in the different clusters; we encode our uncertainty about $S$ with the probability distribution $P(s)$. For each model $s$, we define a continuous vector-valued variable $\Psi_s$, whose instantiations $\psi_s$ correspond to the possible parameters of the model. We encode our uncertainty about $\Psi_s$ with a probability density function $P(\psi_s \mid s)$, as described above.

We score the candidate models by evaluating the *marginal likelihood* of the data set $D$ given the model $s$ [12]. That is, we want to compute

$$P(D \mid s) = \int P(D \mid \psi_s, s) P(\psi_s \mid s) P(s) d\psi_s.$$

The exact computation of the marginal likelihood is intractable for models with hidden variables. We approximate it using a scheme introduced by Cheeseman and Stutz [5]. This approximation is based on the fact that $P(D \mid s)$ can be computed efficiently for complete data. If $D_c$ is any completion of the data set $D$, we have

$$P(D \mid s) = P(D_c \mid s) \frac{\int P(D, \psi_s \mid s) d\psi_s}{\int P(D_c, \psi_s \mid s) d\psi_s}.$$

Letting $\tilde{\psi}_s$ be either an MAP or an ML estimate for $\psi_s$, we can apply the BIC/MDL approximation to the numerator and denominator, and get;

$$\log P(D \mid s) \approx \log P(D_c \mid s) + \log P(D \mid \tilde{\psi}_s, s) - \log P(D_c \mid \tilde{\psi}_s, s).$$

(In our case, the dimension of the complete data is the same as the dimension of the actual data, so the model complexity term cancels out.) We can compute the last two terms in this estimate fairly efficiently by running our EM algorithm from the previous section. Chickering and Heckerman [6] showed that this approximation is surprisingly accurate, much more so than a direct use of BIC/MDL [6].

The first term, $P(D_c \mid s)$, is the probability of a complete data set, where the distribution of the continuous variables in the network, conditioned on each instantiation of the discrete variable *Type*, is a multivariate normal distribution. Geiger and Heckerman [11] show that, in the case of complete data, the marginal likelihood has a closed form that decomposes (as usual) as a product over separate famillies in the model. We omit the (straightforward) details.

Given a scoring function, we can apply standard techniques for finding a high-scoring structure. We use a greedy hill-climbing search with random restarts. Our search space operators modify the subutility structure of each type separately. An operator can add a variable to an existing subutility function, delete a variable from a function, or introduce a new subutility function with a single variable. We evaluate each candidate successor structure by running EM on it, and then scoring it using the Cheeseman-Stutz approximation to the Bayesian score.

## 5 Using the Model for Utility Elicitation

There are many ways to use the model we learn to facilitate utility elicitation and improve the quality of the results.

The most obvious use is simply to use the model as a guide to the range of utility functions within the population. In particular, our model incorporates a built-in measure of confidence. When we assess a new user's utility function, we can immediately discover if he or she is an "outlier" — a person with an atypical utility function. We can ask such a person additional questions to make sure that there was no error in the process.

A somewhat deeper use of the model, along the same lines, is for smoothing the results of the utility elicitation process for a particular individual based on trends in the population as a whole. Given the amount of noise in the utility elicitation process, smoothing of this type is likely to be very useful in getting robust utility estimates.

We can also use the model in a much more fundamental way to change the entire utility elicitation process. For (conditionally) additive decompositions, Keeney and Raiffa [15] describe a utility elicitation procedure which exploits the structure to reduce the number of questions asked. A separate scale is established for every utility function component and the user is asked a series of questions about its parameters. At the end, a new set of assessments must be made to discover the scaling constants. This procedure has become a gold standard in many applications.

This method cannot take advantage of the more generalized factorizations allowed by our algorithm. We propose an alternative procedure which is general enough to handle all factorizations. When we assess the utility function of a new user, we only need to ask as many questions as the number of parameters in our model. The simplest way to choose the outcomes to assess is to convert the projection matrix to the reduced row echelon form and discard the outcomes corresponding to the rows consisting entirely of zeros. Once the values of all the subutility functions are known, we can compute the utility values for the remaining outcomes. It would be good practice to double check that the chosen decomposition really matches the new user's utility function structure by asking a few more "redundant" questions and comparing the answers with those predicted by the function we had computed.

This procedure can also be modified to utilize the model in a more principled way. We can view the utilities elicited for different outcomes as evidence in the distribution defined by the model. We can then use standard probabilistic inference to compute the distribution over the user's subutility functions. The more utilities we elicit, the more evidence we have, the more certain we are about the actual value of the user's subutility functions. We can apply techniques such as conditional mutual information or variance reduction to decide, at each point in time, which utility elicitation question is likely to be the most informative about the subutility variables. We can also make principled decisions on when to stop the elicitation process by considering our uncertainty about these variables.

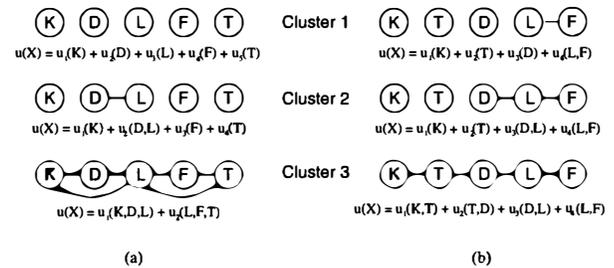

Figure 1: Best decomposition for Visual Analog Scale (a) and Standard Gamble (b).

Finally, we can use probabilistic models of the utility function as the basis for a more targeted process of utility elicitation. In a given decision making task, the utilities of different outcomes typically influence the decision, and the resulting expected utility, to radically different extents. Most simply, some outcomes may have very low probability in the current setting, so their utility is largely irrelevant. Having a distribution over the utility functions in the population, we can compute the value of information of every elicitation question; we can then focus our efforts on those that have the highest impact on our actual decision [4].

## 6 Experimental Results

We tested our approach on both real and synthetically generated data.

Our primary dataset consists of utility functions elicited in a prenatal diagnosis study performed by [17]. All study subjects were recruited from the University of California at San Francisco (UCSF) Prenatal Diagnosis Center. Study subjects were recruited from a counseling session for women who have not yet decided which prenatal diagnostic test to undergo, or, in some cases, whether to undergo prenatal diagnosis at all.

Out of 70 subjects we selected 51 who completed the entire interview, which involved assessing utilities for 22 outcomes using two elicitation methods: standard gamble (SG) and visual analog scale (VAS). These two methods are known to produce very different utility values, thus we treated the two sets of utilities as two distinct databases. We treated the values of all the outcomes the women were not asked about as missing.

We searched the space of 1-, 2- and 3-cluster models. The best models we learned for our two databases were in both cases 3-cluster models. They are presented in Figure 1. The nodes correspond to utility attributes in our domain: testing ($T$), Down's status ($D$), pregnancy loss ($L$), knowledge ($K$) and future pregnancy ($F$). Additive and conditional additive independence corresponds to vertex separation. While the size of the database does not allow us to treat our models as representing the true structure of the utility functions in the population, some of the correlations found are very interesting. For example, the correlation between the utilities for pregnancy loss and utilities for Down's status and future pregnancy are highly intuitive.

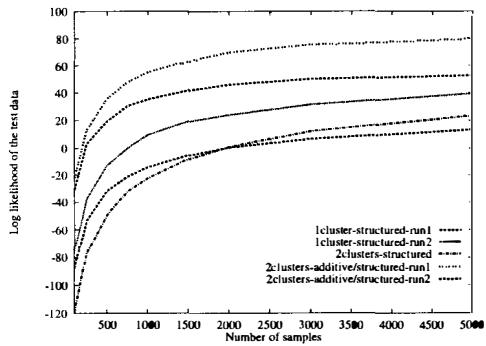

Figure 2: Learning curves for several models.

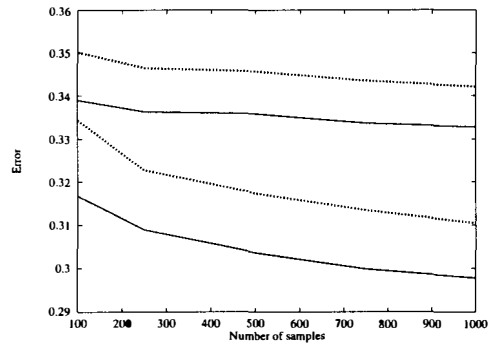

Figure 3: Least-squares projection vs. MAP projection

We note that, in both cases, structures having multiple clusters received substantially higher scores than structures having a single cluster. Furthermore, structures where the different clusters had different decompositions scored more highly than structures where all clusters used the same decomposition. This supports our hypothesis that different subpopulations exist, and have different decompositions.

We also tested our algorithm on synthetic data. In our artificial domain, we had 3 utility attributes, one ternary and two binary, and 12 outcomes. We had three basic structures: fully additive; structured, in which $u(o) = u_1(X_1, X_2) + u_2(X_2, X_3)$; and fully connected (no independencies). We generated 10–20 distributions for each structure, using different parameters.

In one cluster tests, we were always able to recover the structure of the original distribution. For the additive model, the correct structure was chosen after seeing at most 2 data points. (This result was to be expected given the well-known bias towards simpler structures in Bayesian learning.) For the structured model, the number of samples needed ranged from 100 to 750. For the fully connected model, we needed 200-500 samples.

In two-cluster tests, small amounts of data (10–100 samples) always resulted in a model with one fully connected and one fully additive structure, regardless of the underlying distribution. Given more data (1000-5000), we were able to learn either the correct structure or one differing by only one variable's presence or absence in a subutility function. We obtained these results for models with the same as well as with differing decompositions in the different clusters.

We also tested our algorithm as a density estimator. For these tests, we used a domain with 4 attributes, one ternary and three binary. We had two structures: one fully additive and one structured in which $u(o) = u_1(X_1, X_2) + u_2(X_2, X_3) + u_3(X_2, X_4)$. We created several 1- and 2-cluster models, with the same decomposition in different clusters in some models and different decompositions in other models. The learning curve tests are presented in Figure 2. As the number of samples grows, the learned parameters generally seem to converge to the generating distribution.

Finally, we tested the smoothing effect of using parameter priors in our algorithm. After learning the parameters of the model (using the structure our data was generated from), we computed the values of the weight vector $w_t$ using least-squares projection and MAP projection (as described in Section 3) for the samples in our test set. We compared these values to the true weights $w_t$ used to generate these samples. Figure 3 shows the results on 1- (solid lines) and 2-cluster (dotted lines) structured models. The upper curve in both cases corresponds to the least-squares projection, the lower to MAP projection. The error for MAP projection is not only lower, it also decreases more rapidly.

## 7 Conclusion and Extensions

This paper introduces a new approach to acquiring and using preference information. Treating utilities as random variables allows us to deal in a principled way with the uncertainty inherent in utility assessments. It also helps us utilize any prior knowledge we may have.

We have presented an algorithm for learning a probabilistic model of the utility functions in a population of users. Our approach uses Bayesian learning techniques, and utilizes some of the same principles that have been used successfully in structure search for probabilistic models.

Our approach allows us to discover the factorization structure of the utility functions appropriate for a given domain. It accommodates a wide range of possible factorizations, including those corresponding to additive, conditionally additive, and generalized additive independence.

Our approach is significantly more expressive than the naive linear-regression approach in several respects. First, it allows more general notions than simple additive independence; these are far more realistic assumption in many domains. Second, it explicitly accounts for different clusters of users that may use different decompositions. Indeed, our approach discovers interesting structure in the prenatal diagnosis domain of [17], where the traditional linear regression model failed to do so.

The statistical learning perspective also has other benefits. By learning a statistical model of utilities in the population, we are able to associate a "confidence" in our assessment of an individual's utility: if it is extremely unlikely given our model, perhaps fatigue or some other source of noise interfered with the elicitation process. We can also use the model to "smooth" our estimates in a user's utility function, reducing the effects of noise. Finally and most importantly, we can use this statistical model to substantially ease the elicitation process (see [4]).

There are several interesting extensions of this line of work that we would like to pursue. So far, most work (including ours) has focused on notions of independence at the level of variables. In probabilistic settings, this notion has been refined to that of *context-specific independence* [3], which allows independence of two variables $X$ and $Y$ in the context of a particular value $z$ of a third variable $Z$, but not in the context of a value $z'$ for $Z$. An analogous notion can also be defined for utilities. We hope to extend our approach to handle these more refined factorizations of utility functions. In another extension, we hope to capture relations between utility variables and other variables. For example, it has been observed that people who have experienced an outcome tend to assign it a higher utility value than those for whom the outcome is imaginary [18]. This type of correlation can be represented very naturally as a dependence in our probabilistic model; we hope to extend our approach to handle this type of situation.

**Acknowledgments** We would like to thank Ron Parr, Xaver Boyen and Joseph Norman for many useful discussions and Miriam Kuppermann for allowing us to use her data for the prenatal testing domain. We are also grateful to Uri Lerner for his help in building the inference code for Gaussians. This research was supported by ARO under the MURI program "Integrated Approach to Intelligent Systems", grant number DAAH04-96-1-0341 and by ONR contract N66001-97-C-8554 under DARPA's HPKB program.

# A EM Computation

A Normal-Wishart prior defines a distribution over the mean and covariance matrix of a Normal distribution. It is parameterized by: a *precision matrix* $R_t$; a number $\beta_t > m_t - 1$; a mean vector $\lambda_t$; and a number $\nu_t > 0$. Essentially, $R_t$ and $\beta_t$ define a Wishart distribution $w(R_t, \beta_t)$ over $m_t \times m_t$ matrices $Q_t$. The conditional distribution of $\mu_t$ given $Q_t$ is a Gaussian with mean $\lambda_t$ and covariance $\nu_t Q_t^{-1}$.

The conditional distribution of vectors $\mathbf{y}$ given $\mu_t$ and $Q_t$ sampled from this distribution is a Gaussian with mean $\mu_t$ and covariance $v_t Q_t^{-1}$.

The Normal-Wishart distribution is conjugate to the Gaussian distribution. In other words, if we have a Normal-Wishart prior $(R_t^0, \beta_t^0, \lambda_t^0, v_t^0)$, and we observe vectors $\mathbf{y}[1], \ldots, \mathbf{y}[\ell]$ from the associated Gaussian, then the posterior distribution over the parameters is also Normal-Wishart, with the following update rule:

$$\bar{\mathbf{y}} = \frac{1}{\ell} \sum_{j=1}^{\ell} \mathbf{y}[j] \quad (3)$$

$$\lambda_t = \frac{v_t^0 \lambda_t^0 + \ell \bar{\mathbf{y}}}{v_t^0 + \ell} \quad (4)$$

$$v_t = v_t^0 + \ell \quad (5)$$

$$S_t = \sum_{j=1}^{\ell} (\mathbf{y}[j] - \bar{\mathbf{y}})(\mathbf{y}[j] - \bar{\mathbf{y}})^T \quad (6)$$

$$R_t = R_t^0 + S_t + \frac{v_t^0 \ell}{v_t^0 + \ell}(\lambda_t^0 - \bar{\mathbf{y}})(\lambda_t^0 - \bar{\mathbf{y}})^T \quad (7)$$

$$\beta_t = \beta_t^0 + \ell \quad (8)$$

In our setting, we assume that the parameters $\mu_t, \Sigma_t$ of $P(\mathbf{W}_t \mid t)$ are distributed Normal-Wishart with parameters $(R_t^0, \beta_t^0, \lambda_t^0, v_t^0)$. We also assume that the variance $\sigma_t^2$ associated with all of the variables $U_o$ is distributed one-dimensional Wishart with parameters $\rho_t^0, \gamma_t^0$ and $\eta_t^0$. $\rho_t, \gamma_t$ and $\eta_t$ correspond to $R_t, \beta_t$ and $v_t$ in the distribution over $\mathbf{W}_t$ and their update rules are analogous to update rules 7, 8 and 5 respectively.

To do inference with this model, we need to marginalize out the parameter prior and obtain a distribution over the domain variables only. Given a Normal-Wishart parameter distribution $(R_t, \beta_t, \lambda_t, v_t)$, the distribution over $\mathbf{W}_t$ given $t$ is an $n$ dimensional $t$ *distribution*, which can be approximated using a multivariate Gaussian. For the type-specific distributions, we get:

$$\mu_t = \lambda_t$$
$$\Sigma_t = \frac{v_t + 1}{v_t \cdot (\beta_t - m_t - 1)} R_t$$

For the variance $\sigma_t^2$ we set

$$\sigma_t^2 = \frac{\bar{\eta}_t + 1}{\bar{\eta}_t \cdot (\bar{\gamma}_t - 2)} \rho_t.$$

The marginalization for a Dirichlet distribution over the type, with hyperparameters $\alpha_t$, is the standard one: $\theta_t = \alpha_t / (\sum_{t'} \alpha_{t'})$.

When applying EM to our model, the parameters to be estimated are $\theta_t, \mu_t, \Sigma_t$ and $\sigma_t^2$ for every $t$. The hidden variables are $T$ and $\mathbf{W}_t$. In order to complete the data, we must compute $P(T[j], \mathbf{W}_t[j] \mid \mathbf{u}[j], params)$. We marginalize the parameter prior, as we just described. The result is a Gaussian distribution $P(\mathbf{W}_t, \mathbf{U} \mid t)$. For each $t$, we compute $P(\mathbf{W}_t \mid t, \mathbf{u}[j])$ and the marginal $P(\mathbf{u}[j] \mid t)$. We also compute the posterior probability of the different types as $P(t \mid \mathbf{u}[j]) \propto P(t) \cdot P(\mathbf{u}[j] \mid t)$.

Using these probabilities, we can easily compute the (expected) sufficient statistics required for the update of our various parameter priors. For the Dirichlet, we merely need the expected count $\bar{N}(t) = \sum_j P(t \mid \mathbf{u}[j])$. For the various type specific Gaussians, we must compute the expected value of $\lambda_t$ and $S_t$. Intuitively, we have to take the expectation over uncountably many "completed" data cases — a continuum of possible completions for each $j$. Fortunately, this turns out to be easy. The key is that the posterior distribution over $\mathbf{W}_t[j]$ given $t$ is a multivariate Gaussian with mean $\mu_t[j]$ and covariance $\Sigma_t[j]$. Let $\pi_t[j]$ denote $P(t \mid \mathbf{u}[j])$; intuitively $\pi_t[j]$ is the extent to which the $j$th sample belongs to type $t$, and therefore the extent to which it influences the estimate of its parameters. It is straightforward to verify that

$$\bar{\ell} = \sum_{j=1}^{\ell} \pi_t[j]$$

$$\bar{\mathbf{y}}_t = \frac{1}{\bar{\ell}} \sum_{j=1}^{\ell} \pi_t[j] \mu_t[j]$$

$$\bar{S}_t = \sum_{j=1}^{\ell} \pi_t[j] \left( (\mu_t[j] - \bar{\mathbf{y}}_t)(\mu_t[j] - \bar{\mathbf{y}}_t)^T + \Sigma_t[j] \right)$$

Finally, we must compute the expected empirical variance $\bar{s}_t$ needed to update $\rho_t$ and in turn $\sigma_t^2$. Simple linear algebra shows that, if $\mathbf{W}_t$ is distributed Gaussian with mean $\mu_t[j]$ and variance $\Sigma_t[j]$, then $U^* = A_t \mathbf{W}_t$ is distributed Gaussian with mean $A_t \mu_t[j]$ and variance $\Upsilon_t[j] = A_t \Sigma_t[j] A_t^T$. Thus, we get that

$$\bar{s}_t = \sum_{j=1}^{\ell} \pi_t[j] \sum_o (\Upsilon_t(o,o)[j] + ((A_t \mu_t[j])_o - u_o)^2)$$

and

$$\rho_t = \rho_t^0 + \bar{s}_t + \frac{\eta_t^0 n \ell}{\eta_t^0 + n \ell} \cdot \sum_o \sum_{j=1}^{\ell} \pi_t[j]((A_t \mu_t[j])_o - \bar{u}_o)^2.$$

Essentially, the empirical variance has components for different data cases $j$ (which determines $P(\mathbf{W}_t[j] \mid t)$), and outcomes $o$. The contribution for a type $t$ is weighted by its probability. For each $j$ and $o$, there is a contribution for the difference between the mean of $U_o^*$ and the observed utility for outcome $o$, and a contribution for the inherent variance of $U_o^*$.

We can now use these expected sufficient statistics in place of the exact sufficient statistics in Equations (4), (5), (7) and (8). This gives us new estimates of the posterior over the parameters relative to the completed data. We then marginalize the posterior to induce new parameters, and continue.